\documentclass[conference,letterpaper]{IEEEtran}

\usepackage[utf8]{inputenc} 
\usepackage[T1]{fontenc}
\usepackage{url}
\usepackage{ifthen}
\usepackage{cite}
\usepackage[cmex10]{amsmath} % Use the [cmex10] option to ensure complicance
\usepackage{bm}
\usepackage{amsmath}
\usepackage{amssymb}
\usepackage[linesnumbered,ruled]{algorithm2e}
\usepackage{booktabs}
\usepackage{comment}
\usepackage{tcolorbox}
\usepackage{url}
\usepackage{xcolor}
\usepackage{multirow}

\usepackage{breqn,flexisym,array}
\usepackage{graphicx}

\newtheorem{theorem}{Theorem}%[section]
\newtheorem{lemma}{Lemma}%[section]

\newenvironment{lemma-waku}
  {\begin{tcolorbox}[colframe=white, colback=black!5!white, sharp corners=south]%
    \begin{lemma}}%
  {\end{lemma}\end{tcolorbox}}
\newenvironment{theorem-waku}
  {\begin{tcolorbox}[colframe=white, colback=black!5!white, sharp corners=south]%
    \begin{theorem}}%
  {\end{theorem}\end{tcolorbox}}

\newcommand{\ve}{{\bm{e}}}

\newcommand{\x}{{\bm{x}}}

\newcommand{\bE}{{\mathbb{E}}}

\newcommand{\cH}{{\mathcal{H}}}

\newcommand{\cI}{{\mathcal{I}}}

\newcommand{\cL}{{\mathcal{L}}}

\newcommand{\bR}{{\mathbb{R}}}

\newcommand{\cZ}{{\mathcal{Z}}}

\newcommand{\valph}{{\bm{\alpha}}}

\newenvironment{tsaligned}{\begin{equation}\begin{aligned}}{\end{aligned}\end{equation}}
\newenvironment{tsaligned*}{\begin{equation*}\begin{aligned}}{\end{aligned}\end{equation*}}

\newcommand{\tscitelongver}{\textbf{\textcolor{blue}{[arXiv]}}}
\newcommand{\tsqed}{\hfill\ensuremath{\square}}

\newcommand{\tsshort}[1]{{}}
\newcommand{\tslong}[1]{{#1}}

\begin{document}
\title{Linearly Convergent Mixup Learning}
% \title{Mixup Learning with Kernels} 

% %%% Single author, or several authors with same affiliation:
\author{%
\IEEEauthorblockN{Gakuto Obi${}^{\dagger}$, Ayato Saito${}^{\ddagger}$, Yuto Sasaki${}^{\ddagger}$ and Tsuyoshi Kato${}^{\ddagger}$}
\IEEEauthorblockA{${}^{\dagger}$Graduate School of Informatics\qquad ${}^{\ddagger}$Faculty of Informatics\\
                   Gunma University, Japan\\
                   Email: katotsu.cs@gunma-u.ac.jp}% 
}

%%% Several authors with up to three affiliations:
% \author{%
%   \IEEEauthorblockN{Author 1}
%   \IEEEauthorblockA{Department of Electrical Engineering \\
%                     University 1\\
%                     City 1\\
%                     Email: author1@university1.edu}
%   \and
%   \IEEEauthorblockN{Author 2 and Author 3}
%   \IEEEauthorblockA{Research Center XY\\ 
%                     City 2\\
%                     Email: \{author2, author3\}@research-center.com}
% }

\maketitle

%%%%%%
%% Abstract: 
%% If your paper is eligible for the student paper award, please add
%% the comment "THIS PAPER IS ELIGIBLE FOR THE STUDENT PAPER
%% AWARD." as a first line in the abstract. 
%% For the final version of the accepted paper, please do not forget
%% to remove this comment!
%%

\begin{abstract}
    \tsshort{"THIS PAPER IS ELIGIBLE FOR THE STUDENT PAPER AWARD."}
    % Recent advancements in deep learning have highlighted the importance of large-scale data and computational resources, while classical machine learning methods, particularly kernel-based techniques like Support Vector Machines (SVMs) and logistic regression, continue to perform effectively in data-constrained environments. 
    Learning in the reproducing kernel Hilbert space (RKHS) such as the support vector machine has been recognized as a promising technique. It continues to be highly effective and competitive in numerous prediction tasks, particularly in settings where there is a shortage of training data or computational limitations exist.
    These methods are especially valued for their ability to work with small datasets and their interpretability. 
    To address the issue of limited training data, mixup data augmentation, widely used in deep learning, has remained challenging to apply to learning in RKHS due to the generation of intermediate class labels.
    Although gradient descent methods handle these labels effectively, dual optimization approaches are typically not directly applicable. 
    In this study, we present two novel algorithms that extend to a broader range of binary classification models. 
    Unlike gradient-based approaches, our algorithms do not require hyperparameters like learning rates, simplifying their implementation and optimization.
    Both the number of iterations to converge and the computational cost per iteration scale linearly with respect to the dataset size. 
    The numerical experiments demonstrate that our algorithms achieve faster convergence to the optimal solution compared to gradient descent approaches, and that mixup data augmentation consistently improves the predictive performance across various loss functions.
 \end{abstract}
 
\section{Introduction}
% 
% In recent years, deep learning models have made remarkable progress in various fields, although much of this success relies on the use of big data and high-performance computational resources~\cite{hofmann-aos2008}. In contrast, in environments where sufficient training data are unavailable or computational costs are constrained, classical machine learning methods, particularly kernel methods such as SVM and logistic regression, remain effective and continue to be competitive in many prediction tasks~(e.g.~\cite{ZhiyunLu-icassp2016}). 
% Kernel methods are capable of extracting excellent performance even from a small number training examples, and are being re-evaluated in application domains where training examples or computational resources are limited. 
Techniques based on learning in reproducing kernel Hilbert spaces (RKHS), including support vector machines (SVM) and kernel methods, have gained recognition as powerful approaches. These methods remain highly effective and continue to outperform in various prediction tasks, especially in situations where training data is limited or computational resources are restricted~(e.g.~\cite{ZhiyunLu-icassp2016}).

As a method to overcome the issue of insufficient training data, \emph{mixup} data augmentation is often used in the context of deep learning~\cite{HongyiZhang-iclr18}. 
However, the mixup cannot be applied in an uncomplicated manner to some learning algorithms for kernel methods. 
Using mixup augmentation generates intermediate class labels. 
When gradient descent-based optimization algorithms are applied to the primal problem, intermediate class labels do not pose an algorithmic difficulty. 
However, it is not straightforward to apply the conventional optimization algorithms to the dual problem. 
Mochida et al.~\cite{Mochida-ieice24a} found that in the case of SVM, optimization to the dual problem is possible with the classical theories as an exception, even though these theories are not applicable in a direct manner. 

In this paper, we introduce two new algorithms applicable to a broader range of learning models. 
The two new algorithms are based on the stochastic dual coordinate ascent algorithm (SDCA)~\cite{Shalev-Shwartz2013a-SDCA} that is a framework to solve the dual problem. 
Both algorithms inherit the favorable properties of SDCA: 
they do not require any hyperparameters, such as learning rates, and the number of iterations required to converge to the optimal solution and the computational cost per iteration are both guaranteed to scale linearly with respect to the dataset size.
% any hyper-parameters such as learning rates are not required; the number of iterations required to converge to the optimal solution and the computational cost per iteration are both guaranteed to be linear with respect to the dataset size. 
Additionally, numerical experiments confirmed that using these two algorithms developed in this study led to faster convergence to the optimal solution than classical gradient descent-based methods, and that mixup data augmentation improved predictive performance across various loss function. 
\tsshort{%
All the proofs for theorems and lemmas are given in \tscitelongver. %
}% tsshort

\tslong{%
This paper is organized as follows. 
In the next section, we first review the existing research on mixup and highlight the significance of this study. 
Section~\ref{s:primal} formalizes the learning problem in RKHS with the mixup data augmentation, which is the focus of this study. 
Section~\ref{s:naive} demonstrates that the maximization problem does not become trivial when taking the dual problem in a na\"{i}ve way. 
In response, the two solution methods developed in this study are presented in Sections~\ref{s:approx} and \ref{s:newdual}, respectively. 
Section~\ref{s:exp} reports the results of experiments using real-world data, investigating the impact of the mixup data augmentation on the predictive performance. 
Additionally, we demonstrate the efficiency of the two algorithms developed in this study. 
The final section concludes this study.
}% 

\section{Related Work}
\label{s:related}
The success of mixup~\cite{HongyiZhang-iclr18}, a well-known simple data augmentation technique, has led to various extensions and applications across different domains including computer vision~\cite{SangdooYun-2019cutmix,YuxiangSun-ijcnn2023, hendrycks-iclr2020}, facial expression recognition~\cite{Psaroudakis-cvpr2022}, natural language processing~\cite{LichaoSun-iccl2020}, and time series forecasting~\cite{YunZhou-ml4its2023}. 
% Several research provided theoretical insights into the effectiveness of the mixup technique~\cite{Carratino-jmlr2022,YingtianZou-uai2023}. 
% Zhang and Deng~\cite{LinjunZhang-arxiv2021} have shown that by using Taylor expansion, the empirical loss with mixup includes a form of Jacobian regularization. 
% Furthermore, recent literature~\cite{Thulasidasan-neurips2019,Carratino-jmlr2022} suggested the relationship between the mixup and the label smoothing~\cite{Muller-neurips2019}, which involves adding uniform noise to the class labels of the examples. 
Several studies provided theoretical insights into the effectiveness of the mixup technique~\cite{Carratino-jmlr2022,YingtianZou-uai2023,LinjunZhang-arxiv2021,Thulasidasan-neurips2019}. 

% Thus, numerous studies have been conducted on the applications and theoretical analyses of mixup. 
However, there has been little research contributing to optimization algorithms for performing mixup learning in RKHS. 
To the best of our knowledge, the only exception is the work by Mochida et al.~\cite{Mochida-ieice24a}. 
They applied mixup data augmentation to the training of SVM. 
Learning problems in RKHS such as SVM are usually expressed as a convex formulation, and for numerical stability, approaches based on solving the dual problem are generally preferred~\cite{Zishanshao-arxiv2024,QiLei-pmlr17,DejunChu-tnn18,Tran-kdd2015,Takada-ieice24a,TajTsu-icpr21a,TajHir-sdm21a,KatHir-acml19a}. 
However, a technical challenge lies in the dual function that contains the convex conjugate of the mixup loss function. 
This convex conjugate is expressed by an infimal convolution, and generally, computing the value of an infimal convolution requires numerical searching. 
In contrast, Mochida et al.~\cite{Mochida-ieice24a} showed that, in the case of hinge loss used with SVMs, the infimal convolution has a closed form. 
This was an exception in the broader context of learning models. 
When training other kernel methods under the mixup setting, the issue of the convex conjugate has not been addressed so far.  
In this paper, we propose new solutions to address this issue.

\section{Primal problem}
\label{s:primal}
In this section, we formulate the learning problem that arises when the number of examples is increased using the mixup method~\cite{HongyiZhang-iclr18}. 
The mixup method randomly selects two examples from the training dataset, $(\x_{i}^{\text{o}}, y_{i}^{\text{o}})$ and $(\x_{j}^{\text{o}}, y_{j}^{\text{o}}) \in \bR^{d} \times \{\pm 1\}$, and uses a value $\eta \in [0, 1]$ sampled from a Beta distribution. Then, the two examples are combined to create a new example by linearly interpolating both the features and labels:
\begin{tsaligned}
    & \x_{\text{new}} = (1-\eta)\x_{i}^{\text{o}} + \eta\x_{j}^{\text{o}}, \,
    y_{\text{new}} = (1-\eta) y_{i}^{\text{o}} + \eta y_{j}^{\text{o}}.
\end{tsaligned}
This results in a new label that is not strictly binary but lies between the two binary class labels $\pm 1$, which is a key characteristic in the mixup data setting. 
The augmentation process can be repeated to generate additional examples and enhance the training dataset.

% The data augmentation techniques discussed in the study extend beyond Mixup to include its variant, GenLabel. 
% Both methods share the feature of creating intermediate class labels from binary class labels (i.e., $\pm 1$).
% Mixup interpolates the feature vectors and class labels of two randomly chosen examples to generate a new example with an intermediate label, while GenLabel follows a similar principle but may use different mechanisms to generate these intermediate labels.
% The optimization algorithm developed in the study is designed to be flexible and applicable to various data augmentation techniques that produce continuous class labels (ranging from $-1$ to $+1$) from the original binary class labels. 
% This broadens the scope of the algorithm's applicability beyond Mixup and GenLabel to other methods that follow similar principles.

We wish to determine a binary classifier $f: \bR^{d} \to \bR$ from the dataset with continuous class labels
\(
(\x_{1},y_{1}),\dots,(\x_{n},y_{n})\in\bR^{d}\times\bR
\)
generated with some data augmentation method. 

In order to learn a predictor $f$ from mixup data, the mixup loss function defined as 
\begin{tsaligned}\label{eq:assume-phi0}
  \phi_{\text{mup}}(s\,;\,y)
  :=
  \frac{1+y}{2}\phi_{0}(s)
  +
  \frac{1-y}{2}\phi_{0}(-s)
\end{tsaligned}
is used, where the function $\phi_{0}:\bR\to\bR$ is a standard loss function used without intermediate labels (e.g., a binary class label). 
\tsshort{%
Examples of the loss functions~$\phi_{0}$ include the binary cross entropy (BCE) function, the smoothed hinge function, and the quadratic hinge function. Their definitions are given in Appendix of \tscitelongver. %
}% tsshort
\tslong{ %
  Examples of the loss functions~$\phi_{0}$ include the binary cross entropy (BCE) function, the smoothed hinge function, and the quadratic hinge function. Their definitions are given in Appendix~\ref{s:stdloss}. % 
}% tslong
% As described in \eqref{eq:assume-phi0}, 
% the mixup loss function is a mixture of $\phi_{0}(s)$ and $\phi_{0}(-s)$, weighted by the class label $y$. 
Suppose that the loss function $\phi_{0}$ is monotonically decreasing, $\phi_{0}(0)>0$,  and \emph{$1/\gamma_{\text{sm}}$-smooth} with $\gamma_{\text{sm}}>0$~\cite{Urruty01a}, 
and the value of the convex conjugate can be computed in constant time. 

The core idea of kernel methods is to find a function $f$ in the RKHS $\cH_{\kappa}$, where the function $f$ minimizes the regularized empirical risk. 
The function $\kappa:\bR^{d}\times\bR^{d}\to\bR$ is a positive definite kernel such as the RBF kernel and polynomial kernel. 
We employ the standard regularized empirical risk defined as:
\begin{tsaligned}\label{eq:emprisk-def}
 R[f]:= \frac{\lambda}{2}\lVert f \rVert^{2}_{\mathcal{H}_{\kappa}} + \frac{1}{n}\sum_{i=1}^{n}\phi_{\text{mup}}(f(\x_{i})\,;\,y_{i}),
\end{tsaligned}
where $\lambda>0$ is the regularization constant.
The first term 
$\frac{\lambda}{2}\lVert f \rVert^{2}_{\cH_{\kappa}}$ is called the regularization term that helps control the complexity of the model to avoid overfitting, 
where $\lVert f\rVert_{\cH_{\kappa}}$ is the norm of $f$ in the RKHS. 
The second term in \eqref{eq:emprisk-def} measures how well the model $f$ fits the training examples using the mixup loss function.

For more complex models like deep learning models, gradient descent is typically the only feasible optimization method, although it requires careful tuning of step sizes. 
In contrast, kernel methods benefit from optimization algorithms like SDCA~\cite{Shalev-Shwartz2013a-SDCA}, which solve the dual formulation of the problem. 
This is a more efficient method in the case of kernel-based learning as it does not require hyperparameter tuning (like step size adjustment in gradient descent).

When applying the mixup augmentation in RKHS, technical challenges lie in a na\"{i}ve dual problem. 
The straightforward approach may not work efficiently in such a problem in the presence of the mixup data augmentation, which adds complexity to the optimization process.
The next section investigates the details of this issue and explains why optimization becomes more complicated in the mixup data augmentation setting. 

\section{Na\"{i}ve dual problem and its challenge}
\label{s:naive}
If the regularized empirical risk function is convex, solving the dual problem instead of the primal problem is advantageous. One of the advantages is that the dual variables offer a way to avoid storing the infinite-dimensional vector $f$.
The following function is a standard choice for the dual function of the regularized empirical risk: 
\begin{tsaligned}\label{eq:dualobj-def}
D_{0}(\valph)
&
:=
-\frac{\lambda}{2}
\left\lVert 
f_{0,\valph}
\right\rVert_{\cH_{\kappa}}^{2} 
- \frac{1}{n}
\sum_{i=1}^{n}
\phi_{\text{mup}}^{*}(-\alpha_{i}\,;\,y_{i})
\end{tsaligned}
where 
\begin{tsaligned}
    f_{0,\valph} := \frac{1}{\lambda n}\sum_{i=1}^{n}\alpha_{i} \kappa( \x_{i}, \cdot ); 
\end{tsaligned}
Therein, $\phi_{\text{mup}}^{*}(\cdot;y_{i}):\bR\to\bR\cup\{+\infty\}$ represents the convex conjugate of the mixup loss function $\phi_{\text{mup}}(\cdot;y_{i})$. 
If the dual variable vector $\valph_{\star}$ that maximizes the dual function is found successfully, the primal variable optimal to the primal problem is obtained by $f_{\star} = f_{0,\valph_{\star}}$. 

The challenge in solving the dual problem comes from the fact that the convex conjugate of the mixup loss for $i\in[n]$ such that $|y_{i}|<1$ can be expressed as an infimal convolution~\cite{StrombergThomas1996}: 
\begin{tsaligned}
\phi_{\text{mup}}^{*}(-\alpha_{i};y_{i})
=
\inf_{u\in\bR}
\bigg\{
\frac{1+y_{i}}{2}
\phi_{0}^{*}
\left(
\frac{2u}{1+y_{i}}
\right)
\\
+
\frac{1-y_{i}}{2}
\phi_{0}^{*}
\left(
\frac{2(\alpha_{i}+u)}{1-y_{i}}
\right)
\bigg\}. 
\end{tsaligned}
The infimal convolution requires a computationally expensive numerical operation in most cases even if each of $\phi_{0}^{*}$ can be computed with a low computational cost. 
As a result, using this in the dual problem (such as in SDCA) introduces additional complexity. 
In what follows, we shall describe how this convex conjugate complicates optimization when SDCA is applied straightforwardly. 

Consider an iterative algorithm that determines the value of the dual variable vector at the $t$th iteration, denoted by $\valph^{(t)}:=\left[\alpha_{1}^{(t)},\dots,\alpha_{n}^{(t)}\right]^\top$, and let $f^{(t)}:=f_{0,\valph^{(t)}}$. 
Denote the primal and dual objective errors, respectively, by:
\begin{tsaligned}\label{eq:primal-dualobjerr-def}
h_{\text{P}}^{(t)}
:=
R[f^{(t)}]
-
R[f_{\star}], 
\quad
h_{\text{D}}^{(t)}
:=
D_{0}(\valph_{\star})
-
D_{0}(\valph^{(t)}). 
\end{tsaligned}

\begin{lemma-waku}\label{lem:errp-if-geodecr}
Consider a randomized algorithm that computes $\valph^{(t)}\in\bR^{n}$ 
from $\valph^{(t-1)}\in\bR^{n}$. 
Suppose that there exists a constant $\beta$ such that $0 < \beta < 1$ and 
\begin{tsaligned}\label{eq:geodecr}
  &
  \beta\cdot\max\left\{ h_{\text{P}}^{(t-1)}, h_{\text{D}}^{(t-1)} \right\}
  \le
  \bE[h_{\text{D}}^{(t)}] - h_{\text{D}}^{(t-1)}. 
\end{tsaligned}
Then, for any constant $\epsilon_{\text{P}}>0$,  
it holds that $\bE\left[h_{\text{P}}^{(t)}\right]\le \epsilon_{\text{P}}$
for 
\begin{tsaligned}\label{eq:niters-lem:errp-if-geodecr}
    t 
    \ge \frac{1}{\beta}
    \log\left(
    \frac{h_{\text{D}}^{(0)}}{\beta \epsilon_{\text{P}}}
    \right). 
\end{tsaligned}
\end{lemma-waku}
Indeed, SDCA guarantees the inequality~\eqref{eq:geodecr} with $\beta^{-1}=O(n)$, suggesting that the required number of iterations is linear with respect to the dataset size if the logarithmic term is regarded as a constant. 

At each iteration $t$ of SDCA, a single example $i\in[n]$ is selected at random, and the corresponding dual variable $\alpha_{i}$ is updated with the other dual variables $\alpha_{1}, \dots, \alpha_{i-1}$, $\alpha_{i+1}, \dots, \alpha_{n}$ fixed. 
The updated value at the $t$th iteration can be expressed as $\valph^{(t)}=\valph^{(t-1)}+\Delta\alpha_{i}^{(t-1)}\ve_{i}$ where $\Delta\alpha_{i}^{(t-1)}\in\bR$ is the difference from the $i$th dual variable in the previous iteration, and $\ve_{i}$ is the unit vector with $i$th entry one. 
It is ideal to determine the value of $\Delta\alpha_{i}^{(t-1)}$ by maximizing 
\begin{tsaligned}\label{eq:J0-def}
  &
  J^{0}_{t}(\Delta\alpha_{i}^{(t-1)})
  :=
  D_{0}(\valph^{(t-1)}+\Delta\alpha_{i}^{(t-1)}\ve_{i})
  -
  D_{0}(\valph^{(t-1)}). 
\end{tsaligned}
However, this ideal maximization usually complicates computation. 
Instead, the lower bound, say $J^{\text{van}}_{t}(\eta)$, is maximized: for all $\eta\in[0,1]$, 
\begin{tsaligned}\label{eq:J-vanilla-def}
J^{0}_{t}(\eta q_{i,t})
\ge
J^{\text{van}}_{t}(\eta)
  :=
  \frac{\eta F_{\text{van},i}^{(t-1)}}{n}
  +
  \frac{\gamma_{\text{sm}}q_{i,t}^{2}\eta}{2n}
  \left(
    1 - 
    \frac{\eta}{\bar{s}_{t}}
  \right) 
\end{tsaligned} 
where $z_{i}^{(t-1)} := f^{(t-1)}(\x_{i})$, 
$u_{i}^{(t-1)} := -\nabla\phi_{\text{mup}}(z_{i}^{(t-1)}\,;\,y_{i})$,
\begin{tsaligned}\label{eq:vanilla-u-q-F-K-sbar}
    &
\quad
q_{i,t} := u_{i}^{(t-1)} - \alpha_{i}^{(t-1)},
\quad
K_{i,i}
:=
\kappa(\x_{i},\x_{i}), 
\\     
& 
F_{\text{van},i}^{(t-1)} 
:= 
\phi_{\text{mup}}(z_{i}^{(t-1)}\,;\,y_{i})
+
\phi_{\text{mup}}^{*}(-\alpha_{i}^{(t-1)}\,;\,y_{i})
\\
&
\qquad 
+
\alpha_{i}^{(t-1)}z_{i}^{(t-1)}, 
\quad
\bar{s}_{t}
:=
\frac{\lambda n \gamma_{\text{sm}}}{K_{i,i}+\lambda n \gamma_{\text{sm}}}. 
\end{tsaligned}
Therein, $\nabla\phi_{\text{mup}}(z\,;\,y)$ denotes the derivative of $\phi_{\text{mup}}(z\,;\,y)$ with respect to $z$. 
The lower bound forms a simple parabola, making the line search quite simple: 
\begin{tsaligned}\label{eq:eta-vanilla}
    \eta_{\text{van},t}
    :=
    \min\left\{1,
    \bar{s}_{t}
    \cdot
    \frac{F_{\text{van},i}^{(t-1)}+\frac{1}{2}\gamma_{\text{sm}} q_{i,t}^{2}}{\gamma_{\text{sm}} q_{i,t}^{2}}
    \right\}. 
\end{tsaligned}
Thus, the vanilla SDCA includes the update rule that determines the step size optimal to the introduced parabola function. This eliminates the need for hyperparameter tuning, enabling numerically stable optimization. 

\begin{algorithm}[t!]
\caption{
  $\textsc{MixupSDCA}_{\text{na\"{i}ve}}$. % for maximizing $D_{0}(\valph)$.
\label{algo:sdca-vanilla}}
\Begin{
Choose $\valph^{(0)}\in\text{dom}(-D_{0})$\; 
$f^{(0)}:= \frac{1}{\lambda n}\sum_{i=1}^{n}\alpha_{i}^{(0)}\kappa(\x_{i},\cdot)$\; 
\For{$t:=1$ \KwTo $T$}{
Select $i$ at random from $\{1,\dots,n\}$\;
$z_{i}^{(t-1)} := f^{(t-1)}(\x_{i})$\;
Compute $F_{\text{van},i}^{(t-1)}$ and $\bar{s}_{t}$ by \eqref{eq:vanilla-u-q-F-K-sbar}\; 
Use \eqref{eq:eta-vanilla} to compute ${\eta}_{\text{van},t}$\;
$\valph^{(t)}:=\valph^{(t-1)}+q_{i,t}{\eta}_{\text{van},t}\ve_{i}$\;
$f^{(t)}:=f^{(t-1)} + \frac{1}{\lambda n}\kappa(\x_{i},\dot)q_{i,t}{\eta}_{\text{van},t}$\; 
}
}
\end{algorithm}
The vanilla SDCA applied to maximizing $D_{0}(\valph)$ is summarized in Algorithm~\ref{algo:sdca-vanilla}. 
When this update rule is implemented in practice, the dual objective error decreases geometrically with $\beta^{-1}=O(n)$, ensuring the linear convergence to the minimum. 
However, there is a significant obstacle to executing this update rule.
The variable $F_{\text{van},i}^{(t-1)}$ involves the convex conjugate of the mixup loss function, and its computation requires a numerical search method, as previously discussed. 
To overcome this obstacle, the authors have identified two approaches.
Each approach shall be introduced in a separate section that follows. 
\section{Approximation approach}
\label{s:approx}
As discussed in the previous section, handling the infimal convolutions poses a challenge for maximizing the dual function $D_{0}(\valph)$. 
The infimal convolution requires numerical search to compute the variable $F_{\text{van},i}^{(t-1)}$, which appears in the step-size update rule of the vanilla SDCA. 
This section presents a solution by approximating the computation of $F_{\text{van},i}^{(t-1)}$. 
The algorithm introduced in this section is capable of performing this approximation while guaranteeing the existence of the coefficient~$\beta$ satisfying \eqref{eq:geodecr}. 

\tslong{%
The approximation approach is described in Algorithm~\ref{algo:sdca-approx}. 
}%
As explained in the previous section, the step size in vanilla SDCA is determined by the vertex of the parabola $J^{\text{van}}_{t}(\eta)$, where $F_{\text{van},i}^{(t-1)}$ is the coefficient of that parabola. 
Replacing the coefficient $F_{\text{van},i}^{(t-1)}$ with a smaller value $\widetilde{F}_{t}$ does not exceed the parabola for any $\eta\in[0,1]$. 
The approximation algorithm introduced here finds a lower value $\widetilde{F}_{t}$ that closely approximates $F_{\text{van},i}^{(t-1)}$ while maintaining the property of linear convergence. 

% Assume that $\phi_{0}^{*}(-\alpha)$ can be computed with $O(1)$ cost. Then, 
For the case of $y_{i}\in\{\pm 1\}$, it is straightforward to compute the value of $F_{i}^{(t-1)}$. 
Let $\widetilde{F}_{t}:=F_{i}^{(t-1)}$ in this case. 
Hereinafter, we shall focus on the case that $|y_{i}|<1$. 
% Algorithm~\ref{algo:find-tilF} can find $\widetilde{F}_{t}\in\bR$ such that 
% the inequality $\widetilde{F}_{t} \le F_{i}^{(t-1)}$ is ensured.  
% 
\begin{algorithm}[t!]
  \caption{
    Determine $\widetilde{F}_{t}$. 
    \label{algo:find-tilF}}
  \Begin{
    Let $\alpha_{i}^{\diamond}:=-\nabla\phi_{\text{mup}}(0\,;\,y_{i})$\; 
    \eIf{$\alpha_{i}^{\diamond}<\alpha_{i}^{(t-1)}$}{
      Find $\widetilde{\zeta}_{t}\in\bR$ such that
      \(
         \widetilde{\zeta}_{t}\le 0 \text{ and }-\nabla\phi_{\text{mup}}(\widetilde{\zeta}_{t}\,;\,y_{i}) \le \alpha_{i}^{(t-1)}
      \)
      \label{line:zetat-right}
    }{
      Find $\widetilde{\zeta}_{t}\in\bR$ such that
      \(
        \widetilde{\zeta}_{t}\ge 0 \text{ and }-\nabla\phi_{\text{mup}}(\widetilde{\zeta}_{t}\,;\,y_{i}) \ge \alpha_{i}^{(t-1)}
        \label{line:zetat-left}
      \)
    }
    Let $\widetilde{\alpha} := -\nabla\phi_{\text{mup}}(\widetilde{\zeta}_{t}\,;\,y_{i})$; 
    Use \eqref{eq:compute-tilF} to compute $\widetilde{F}_{t}$\;
  }
\end{algorithm}
In this case, the value of $\widetilde{F}_{t}$ is determined by 
\begin{tsaligned}\label{eq:compute-tilF}
  \widetilde{F}_{t}
= 
\phi_{\text{mup}}(z_{i}^{(t-1)}\,;\,y_{i})
-
\widetilde{\alpha}\widetilde{\zeta}_{t}
+
\alpha_{i}^{(t-1)}z_{i}^{(t-1)}
-
\phi_{\text{mup}}(\widetilde{\zeta}_{t}\,;\,y_{i})
\end{tsaligned}
where $\widetilde{\alpha}$ and $\widetilde{\zeta}_{t}$ are computed in Algorithm~\ref{algo:find-tilF}. 
These $\widetilde{\zeta}_{t}\in\bR$ and $\widetilde{\alpha}\in\bR$ satisfy
\begin{tsaligned}
  \widetilde{\zeta}_{t} = -\nabla\phi^{*}_{i}(-\widetilde{\alpha})
  \text{ and }
  \phi^{*}_{i}(-\widetilde{\alpha})
   \le \phi^{*}_{i}(-\alpha^{(t-1)}_{i}). 
\end{tsaligned}   
The following lemma is used to prove the linear convergence of this approximation approach. 
\begin{lemma-waku}\label{lem:tilF-le-F}
  Algorithm~\ref{algo:find-tilF} ensures the inequality: 
  $\widetilde{F}_{t} \le F_{i}^{(t-1)}$. 
\end{lemma-waku}

% Hereinafter, we assume that
% \begin{tsaligned}
%   & \forall i\in[n], \quad
%   \exists \rho_{i}\in[0,1], 
%   \\
%   &
%   \phi_{\text{mup}}(z) = \rho_{i}\phi_{0}(z) + (1-\rho_{i})\phi_{0}(-z), 
%   \\
%   &
%   \phi_{0}(0) > 0, \qquad \lim_{z\to-\infty}\nabla \phi_{0}(z) < 0. 
% \end{tsaligned}
%
\newcommand{\ndiv}{{n}}
\newcommand{\czeta}{{4}}

Our implementation of Line~\ref{line:zetat-right} and Line~\ref{line:zetat-left} that determine the value of $\zeta_{t}$ is described as follows. 
Define 
\begin{tsaligned}
  & \cL_{i}
  :=
  \left\{
  \zeta \in\bR
  \,\middle|\,
  \phi_{\text{mup}}(\zeta\,;\,y_{i}) \le n \phi_{0}(0)
  \right\}, 
  \\
& 
b_{u,i} := \inf\cL_{i} \text{ and }
b_{\ell,i} := \sup\cL_{i}. 
\end{tsaligned}
Let $\alpha_{i}^{\diamond}:=-\nabla\phi_{\text{mup}}(0\,;\,y_{i})$
as defined in Algorithm~\ref{algo:find-tilF}. 
There are two cases:
(i) $\alpha_{i}^{(t-1)} < \alpha_{i}^{\diamond}$
and 
(ii)
$\alpha_{i}^{\diamond} \le \alpha_{i}^{(t-1)}$. 
For the first case, 
the value of $\widetilde{\zeta}_{t}$ is determined by
\begin{tsaligned}\label{eq:zetat-when-alph-larger}
  & \widetilde{\zeta}_{t} := \min\bigg\{
    \zeta\in\overline{\cZ}_{\text{p},i}
    \,\bigg|\,
    \alpha_{i}^{(t-1)}\le -\nabla\phi_{\text{mup}}(\zeta\,;\,y_{i}) 
  \bigg\}
\end{tsaligned}
where 
\begin{tsaligned}
  \overline{\cZ}_{\text{p},i}
  :=
  \bigg\{
    \exp\left(
      \frac{k}{\ndiv}
      \left( \czeta + \log b_{\ell,i}\right) - \czeta
    \right)
    \,\bigg| \,
    k\in\{0\}\cup [\ndiv]\bigg\}. 
\end{tsaligned}
For the second case (i.e. 
(ii)
$\alpha_{i}^{\diamond} \le \alpha_{i}^{(t-1)}$),  
the value of $\widetilde{\zeta}_{t}$ is set to
\begin{tsaligned}\label{eq:zetat-when-alph-smaller}
  & \widetilde{\zeta}_{t} := \max\bigg\{
    \zeta\in\overline{\cZ}_{\text{n},i}
    \,\bigg|\,
    -\nabla\phi_{\text{mup}}(\zeta\,;\,y_{i}) \le \alpha_{i}^{(t-1)}
  \bigg\}
\end{tsaligned}
where 
\begin{tsaligned}
  \overline{\cZ}_{\text{n},i}
  :=
  \bigg\{
    -\exp\left(
      \frac{k}{\ndiv}
      \left( \czeta + \log(-b_{u,i})\right) - \czeta
    \right)
    \,\bigg| \,
    k\in\{0\}\cup [\ndiv]\bigg\}. 
  \end{tsaligned}
%
% Therein, $c_{\text{div}}$ is a positive constant number, and $n_{\text{div}}\in\bN$. 
%
The step size for this approximation approach is determined by 
\begin{tsaligned}
  \eta_{\text{approx},t}
  :=
  \min\left\{1,
  \bar{s}_{t}\cdot
  \max\left\{
    1, 
  \frac{\widetilde{F}_{t}+\frac{1}{2}\gamma_{\text{sm}} q_{i,t}^{2}}{\gamma_{\text{sm}} q_{i,t}^{2}}
  \right\}\right\}.  
\end{tsaligned}
With this implementation for Line~\ref{line:zetat-right} and Line~\ref{line:zetat-left},  Algorithm~\ref{algo:find-tilF} ensures that the time taken for each iteration is proportional to the size of the dataset, maintaining linear scalability. 

\tslong{% 
\begin{algorithm}[t!]
  \caption{
    $\textsc{MixupSDCA}_{\text{approx}}$ for maximizing $D_{0}(\valph)$.
  \label{algo:sdca-approx}}
  \Begin{
  Choose $\valph^{(0)}\in\text{dom}(-D_{0})$\; 
  $f^{(0)}:= \frac{1}{\lambda n}\sum_{i=1}^{n}\alpha_{i}^{(0)}\kappa(\x_{i},\cdot)$\; 
  \For{$t:=1$ \KwTo $T$}{
  Select $i$ at random from $\{1,\dots,n\}$\;
  $z_{i}^{(t-1)} := f^{(t-1)}(\x_{i})$\;
  Compute $\bar{s}_{t}$ by \eqref{eq:vanilla-u-q-F-K-sbar}\; 
  Run Algorithm~\ref{algo:find-tilF} to compute $\widetilde{F}_{t}$\; 
  Use \eqref{eq:eta-vanilla} to compute ${\eta}_{\text{van},t}$\;
  $\valph^{(t)}:=\valph^{(t-1)}+q_{i,t}{\eta}_{\text{van},t}\ve_{i}$\;
  $f^{(t)}:=f^{(t-1)} + \frac{1}{\lambda n}\kappa(\x_{i},\dot)q_{i,t}{\eta}_{\text{van},t}$\; 
  }
  }
\end{algorithm}
}% 
% \begin{lemma-waku}\label{lem:zetat-needs-linear-time}
%   Assume that $\widetilde{\zeta}_{t}$ is computed by \eqref{eq:zetat-when-alph-larger} if $\alpha_{i}^{\diamond} < \alpha^{(t-1)}_{i}$; otherwise, \eqref{eq:zetat-when-alph-smaller} is used. 
%   Then, computation of $\widetilde{\zeta}_{t}$ is done with runtime cost $O(\ndiv)$. 
% \end{lemma-waku}
%
The approximation approach also guarantees that the number of iterations scales in a linear fashion with respect to the dataset size.

\begin{theorem-waku}\label{thm:approx-beta}
  Let $R_{\text{mx}}:=\max_{i\in[n]}\sqrt{K_{i,i}}$. 
  Using the approximation algorithm, 
  $\bE\left[h_{\text{P}}^{(t)}\right]\le \epsilon_{\text{P}}$ holds 
  for iteration $t$ such that \eqref{eq:niters-lem:errp-if-geodecr} is satisfied by
  \begin{tsaligned}
   \beta^{-1} = n + \frac{R^{2}_{\text{mx}}}{\lambda \gamma_{\text{sm}}}. 
  \end{tsaligned}
\end{theorem-waku}
\section{Decomposition approach}
\label{s:newdual}
This section provides a second approach to tackle the issues caused by the mixup loss in the dual problem. 
The dual function in \eqref{eq:dualobj-def}, denoted as $D_{0}$, is not unique in general: for a given primal objective function, there can be an infinite number of corresponding dual functions. 
Choosing which dual function to use impacts the ease of optimization. 
This section introduces a new dual function $\widetilde{D}$ that is an alternative choice to $D_{0}$. 
The new dual function $\widetilde{D}$ is obtained by rearranging the regularized empirical risk: 
\begin{tsaligned}\label{eq:emprisk-rearranged}
R[f] = \frac{\lambda}{2}\lVert f \rVert_{\cH_{\kappa}}^{2} + \frac{1}{\widetilde{n}}\sum_{i\in\cI}\widetilde{\phi}_{i}(\sigma_{i} f(\x_{i}))
\end{tsaligned}
where 
\begin{tsaligned}
    & 
\forall i\in[n], \quad \sigma_{i} := 1, \quad \sigma_{i+n} := -1, 
\quad y_{i+n} := -y_{i},  
\\
&
\x_{i+n} := \x_{i}, \quad
\cI :=
\left\{ i\in[2n] \,\middle|\,1+y_{i} > 0
\right\}, \quad \widetilde{n} := |\cI|, 
\\
&
\forall i\in\cI, \quad \widetilde{\phi}_{i} := (1+y_{i})\phi_{0}. 
\end{tsaligned}
Equation \eqref{eq:emprisk-rearranged} is derived by substituting the definition of the loss function into the regularized empirical risk.
In \eqref{eq:emprisk-rearranged}, the losses are decomposed into at most two per example. 
Notice that this reformulation retains a form of a typical regularized empirical risk minimization problem, where the loss for $\widetilde{n}$ training examples 
$(\x_{i},\sigma_{i})_{i\in\cI}\subseteq \bR^{d}\times\{\pm 1\}$ 
is evaluated with $\widetilde{\phi}_{i}$, and no intermediate labels are involved. 
The dual function to \eqref{eq:emprisk-rearranged} is given by 
\begin{tsaligned}\label{eq:dualobj-def-til}
   \widetilde{D}(\valph)
   &
   :=
   -\frac{\lambda}{2}
   \left\lVert 
      \widetilde{f}_{\valph}
   \right\rVert_{\cH_{\kappa}}^{2} 
   -
   \frac{1}{\widetilde{n}}\sum_{j\in\cI}\widetilde{\phi}^{*}_{j}(-\alpha_{j}).    
\end{tsaligned}
where 
\begin{tsaligned}
   \widetilde{f}_{\valph}
   :=
   \frac{1}{\lambda\widetilde{n}}\sum_{j\in\cI}\alpha_{j}\sigma_{j}\kappa(\x_{j (\text{mod}\,n)},\cdot) 
\end{tsaligned}
The smoothness of $\widetilde{\phi}_{i}:\bR\to\bR$ affects the convergence speed of this decomposition approach. 
\begin{lemma-waku}\label{lem:tilphi-smooth}
%
% If $\phi_{0}$ is a $1/\gamma_{\text{sm}}$-smooth function, then 
Function $\widetilde{\phi}_{i}$ is $2/\gamma_{\text{sm}}$-smooth, 
and its convex conjugate can be computed in constant time. 
\end{lemma-waku}
This property ensures the efficiency of direct application of the vanilla SDCA to maximizing the alternative dual function $\widetilde{D}$ without the approximation approach described in the previous section. 

\begin{theorem-waku}\label{thm:decomp-beta}
   Let $R_{\text{mx}}:=\max_{i\in[n]}\sqrt{K_{i,i}}$. 
If we apply the vanilla SDCA to maximizing $\widetilde{D}(\valph)$, 
then the inequality 
$\bE\left[h_{\text{P}}^{(t)}\right]\le \epsilon_{\text{P}}$ holds 
for iteration $t$ such that \eqref{eq:niters-lem:errp-if-geodecr} is satisfied by
\begin{tsaligned}
   \beta^{-1} = 2n + \frac{2R^{2}_{\text{mx}}}{\lambda \gamma_{\text{sm}}}. 
\end{tsaligned}
\end{theorem-waku}

\begin{table}[ht]
\centering
\small
\caption{AUROC for toxity prediction. \label{tab:auroc}}
\begin{tabular}{|l|cc|cc|}
    \hline
    & 
    \multicolumn{2}{|c|}{Neurotoxicity}
    &
    \multicolumn{2}{|c|}{Cardiotoxicity}
    \\
    & Classical & Mixup & Classical & Mixup \\
    \hline
    Smoothed hinge loss & $ 0.881 $ & $ 0.909 $ 
    & $0.818$ & $ 0.857 $  \\
    BCE loss       & $0.888$   & $\mathbf{0.923}$ 
    & $0.832$ & $\mathbf{0.902}$ \\
    Quadratic hinge loss & $ 0.685 $ & $ 0.797 $ 
    & $0.643$ & $ 0.862 $  \\
    \hline
\end{tabular}
\end{table}
\begin{table}[ht]
\centering
\small
\caption{Runtimes until the primal objective error reaches below a threshold of $10^{-5}$. The unit is seconds. \label{tab:runtime}}
\begin{tabular}{|l|ccc|}
    \hline
    & magic04 & bank & spambase \\    
    \hline
    $\textsc{MixupSDCA}_{\text{na\"{i}ve}}$& $7.25\cdot 10^{2}$& $2.36\cdot 10^{2}$& $4.36\cdot 10^{2}$ \\
    $\textsc{MixupSDCA}_{\text{approx}}$& 
    $\bm{2.52\cdot 10^{2}}$& 
    $\bm{1.02\cdot 10^{2}}$& 
    $\bm{1.24\cdot 10^{2}}$ \\
    $\textsc{MixupSDCA}_{\text{decomp}}$& $5.98\cdot 10^{2}$& $2.32\cdot 10^{2}$& $2.57\cdot 10^{2}$ \\
    SGD with $\eta=10^{-1}$& N/A& N/A& N/A \\
    SGD with $\eta=10^{-2}$& N/A& $1.35\cdot 10^{3}$& $3.53\cdot 10^{2}$ \\
    SGD with $\eta=10^{-3}$& $2.69\cdot 10^{2}$& $4.51\cdot 10^{2}$& N/A \\
    SGD with $\eta=10^{-4}$& N/A& N/A& N/A \\
    \hline
\end{tabular}
\end{table}
    
\section{Experiments}
\label{s:exp}

\subsection{Prediction performance}

% In the context of toxicity prediction using pluripotent stem cells, we investigated whether data augmentation via Mixup is effective for various kernel methods. Changes in gene expression levels resulting from exposing pluripotent stem cells, such as embryonic stem (ES) cells, to chemicals are thought to reflect the physiological activity of those chemicals in the human body. Therefore, predicting toxicity based on these gene expression changes has gained attention as a promising approach for evaluating toxicity risk. This method offers the advantage of assessing comprehensive and long-term toxicity risks with a single experiment, significantly reducing the cost of chemical safety evaluations and shortening testing periods~\cite{YamAbuIma16a,Yamane-iScience22a}.

We examined the predictive performance for mixup learning on toxicity prediction from gene expressions~\cite{YamAbuIma16a,Yamane-iScience22a}. 
The experimental conditions followed the methodology outlined in \cite{Mochida-ieice24a}. Specifically, we used gene expression data obtained by exposing stem cells to 24 different chemicals, as used by Yamane et al.~\cite{Yamane-iScience22a}. The neurotoxicity and cardiotoxicity statuses of these 24 chemicals are already known. Based on this, we prepared two benchmark datasets for binary classification. For neurotoxicity, there were 13 positive chemicals and 11 negative chemicals, while for cardiotoxicity, there were 15 positive chemicals and nine negative chemicals.

% The feature vectors were constructed according to previous research~\cite{Yamane-iScience22a,Mochida-ieice24a} as follows: genes were selected based on their contribution to the principal components, and for each chemical, 20 gene networks were constructed. The adjacency matrices of these 20 gene networks were vectorized to form the features for an SVM model, and the prediction scores were then used as the final features. This resulted in feature vectors $\x_{i}^{\text{o}}\in\bR^{20}$ and binary class labels $y_{i}^{\text{o}}\in\{\pm 1\}$ for each chemical.

Prediction performance was evaluated using leave-one-out cross-validation, where one of the 24 chemical substances was used as the test set, and the remaining 23 were used for training.
The 23 training examples were augmented using the mixup data augmentation, increasing the number of examples by 50. Three types of loss functions were employed: smoothed hinge loss, BCE loss, and squared hinge loss. The RBF kernel was used for the kernel function. The regularization parameter $\lambda$ and the kernel width of the RBF kernel were optimized by performing another leave-one-out cross-validation on the 23 training examples. 

The prediction performance was assessed using the Area Under the Receiver Operating Characteristic (AUROC) curve. 
Since the data augmentation using mixup involves randomization, the results vary across trials. Therefore, the average AUROC from five trials was used for evaluation.  
The prediction performance for neurotoxicity and cardiotoxicity is presented in Table~\ref{tab:auroc}. For both types of toxicity, and for all loss functions used, mixup was found to improve the prediction performance. For both neurotoxicity and cardiotoxicity, BCE loss achieved higher AUROC values than smoothed hinge loss. These results suggest that mixup enhances generalization across various machine learning models. 
% , and that SVM is not necessarily the optimal model for this context.

\subsection{Runtime}

To evaluate the convergence speed of the new algorithms to the optimal solution, we conducted numerical experiments. %  using the Mixup data from the toxicity prediction experiment. 
We here denote by $\textsc{MixupSDCA}_{\text{na\"{i}ve}}$, $\textsc{MixupSDCA}_{\text{approx}}$ and $\textsc{MixupSDCA}_{\text{decomp}}$ the algorithms presented in Sections~\ref{s:naive},~\ref{s:approx} and \ref{s:newdual}, respectively. 
We compared the two new algorithms with the stochastic gradient descent (SGD). Four step sizes for SGD were tested: $\eta = 10^{-4}$, $10^{-3}$, $10^{-2}$, $10^{-1}$. 
We employed BCE as the loss function. 
% The goal of this experiments was to examine how quickly the algorithms converged to the minimum by evaluating the primal objective value $R[f^{(t)}]$ during the iterations. 
% The primal objective error was used as a measure of convergence, despite not knowing the true minimum of the primal objective function. 
% To approximate the primal objective error, we ran the new algorithm to compute the dual objective at $T^{\prime}:=10^{6}n$ iterations, and the gap between the primal objective value $R[f^{(t)}]$ and the dual objective value at $T^{\prime}$th iteration, say $R[f^{(t)}]-D_{0}(\valph^{T^{\prime}})$ was regarded as the primal objective error. 
% In fact, we verified that the duality gap at the $T^{\prime}$th iteration satisfies $R[f^{(T^{\prime})}] - D_{0}(\valph^{(T^{\prime})}) < 10^{-10}$. 
% This implies that $0 \le R[f_{\star}] - D_{0}(\valph^{(T^{\prime})}) < 10^{-10}$, meaning that the approximation of $D_{0}(\valph^{(T^{\prime})})$ as the minimal primal objective value is highly accurate and reliable. 
The datasets used in these experiments include magic04, bank, and spambase. 
Each dataset contains 10,000, 4,532, and 4,601 examples, respectively.
For each dataset, 5,000 examples were added using the mixup data augmentation. 
Four values of the regularization parameter were explored ($\lambda=10^{0}/n, 10^{-1}/n, 10^{-2}/n$) and the total runtime was measured. 
If the primal objective error did not converge to $10^{-5}$ within 5,000 epochs, the optimization algorithm was terminated. 

Table~\ref{tab:runtime} displays the runtimes. 
Therein, N/A indicates that the primal objective error did not reach $10^{-5}$ for any setting of the regularization parameter within 5,000 epochs.
The approximation approach converged the most rapidly across all datasets.
For the three datasets, the decomposition approach took 2.37, 2.22, 2.07 times longer than the approximation approach. 
As proven in Theorem~\ref{thm:decomp-beta}, the increase in the required number of iterations led to this result. 
Solving the na\"{i}ve dual problem was slower than the approximation approach due to its dependence on numerical search for computing the convex conjugate. 
For SGD, the step size that completed convergence within 5,000 epochs varied depending on the dataset. 
For the bank dataset, learning converged within 5,000 epochs when $\eta=10^{-2}$ and $\eta=10^{-3}$, but for other step sizes, the primal objective error failed to reach $10^{-5}$ within the 5,000 epochs.
For the magic04 and spambase datasets, learning was completed within 5,000 epochs only when $\eta=10^{-3}$ and $\eta=10^{-2}$, respectively.
For any step size of SGD, the runtime was slower than that of the approximation approach.

\section{Conclusions}
\label{s:concl}

This paper investigates the optimization problem of learning from a mixup-augmented dataset in RKHS and presents two novel algorithms: the approximation method and the decomposition method. 
Both algorithms are guaranteed to converge linearly to the minimum value.
We empirically showed that mixup improves generalization performance in learning binary classification problems on RKHS.
Furthermore, numerical experiments demonstrated that the approximation method converges more rapidly than gradient descent-based approaches and the na\"{i}ve algorithm.

While this paper primarily addresses binary classification, future work remains to analyze the optimization of mixup data augmentation in other machine learning tasks. 
Moreover, the application of mixup learning to federated learning, especially extensions that strengthen privacy and security, is an intriguing direction. 

\section*{Acknowledgment}
This research was supported by JSPS KAKENHI Grants 22K04372 and 23K28245
and JST ASTEP JPMJTR234E. 

\bibliographystyle{IEEEtran}

\begin{thebibliography}{10}
\providecommand{\url}[1]{#1}
\csname url@samestyle\endcsname
\providecommand{\newblock}{\relax}
\providecommand{\bibinfo}[2]{#2}
\providecommand{\BIBentrySTDinterwordspacing}{\spaceskip=0pt\relax}
\providecommand{\BIBentryALTinterwordstretchfactor}{4}
\providecommand{\BIBentryALTinterwordspacing}{\spaceskip=\fontdimen2\font plus
\BIBentryALTinterwordstretchfactor\fontdimen3\font minus
  \fontdimen4\font\relax}
\providecommand{\BIBforeignlanguage}[2]{{%
\expandafter\ifx\csname l@#1\endcsname\relax
\typeout{** WARNING: IEEEtran.bst: No hyphenation pattern has been}%
\typeout{** loaded for the language `#1'. Using the pattern for}%
\typeout{** the default language instead.}%
\else
\language=\csname l@#1\endcsname
\fi
#2}}
\providecommand{\BIBdecl}{\relax}
\BIBdecl

\bibitem{ZhiyunLu-icassp2016}
Z.~Lu, D.~Quo, A.~B. Garakani, K.~Liu, A.~May, A.~Bellet, L.~Fan, M.~Collins,
  B.~Kingsbury, M.~Picheny, and F.~Sha, ``A comparison between deep neural nets
  and kernel acoustic models for speech recognition,'' in \emph{2016 IEEE
  International Conference on Acoustics, Speech and Signal Processing
  (ICASSP)}.\hskip 1em plus 0.5em minus 0.4em\relax IEEE, Mar. 2016, pp.
  5070–--5074, doi: 10.1109/ICASSP.2016.7472643.

\bibitem{HongyiZhang-iclr18}
H.~Zhang, M.~Cisse, Y.~N. Dauphin, and D.~Lopez-Paz, ``mixup: Beyond empirical
  risk minimization,'' in \emph{International Conference on Learning
  Representations}, 2018.

\bibitem{Mochida-ieice24a}
R.~Mochida, M.~Nakajima, H.~Ono, T.~Ando, and T.~Kato, ``Mixup svm learning for
  compound toxicity prediction using human pluripotent stem cells,''
  \emph{IEICE Transactions on Information and Systems}, vol. E107.D, no.~12,
  pp. 1542–--1545, Dec. 2024, doi: 10.1587/transinf.2024edl8040.

\bibitem{Shalev-Shwartz2013a-SDCA}
S.~Shalev-Shwartz and T.~Zhang, ``Stochastic dual coordinate ascent methods for
  regularized loss,'' \emph{J. Mach. Learn. Res.}, vol.~14, no.~1, pp.
  567--599, Feb. 2013.

\bibitem{SangdooYun-2019cutmix}
S.~Yun, D.~Han, S.~J. Oh, S.~Chun, J.~Choe, and Y.~Yoo, ``Cutmix:
  Regularization strategy to train strong classifiers with localizable
  features,'' \emph{arXiv: 1905.04899}, 2019.

\bibitem{YuxiangSun-ijcnn2023}
Y.~Sun, K.~Qi, Y.~Zhou, and Y.~Qi, ``Strip-cutmix for person
  re-identification,'' in \emph{2023 International Joint Conference on Neural
  Networks (IJCNN)}.\hskip 1em plus 0.5em minus 0.4em\relax IEEE, Jun. 2023,
  pp. 1–--8.

\bibitem{hendrycks-iclr2020}
D.~Hendrycks, N.~Mu, E.~D. Cubuk, B.~Zoph, J.~Gilmer, and B.~Lakshminarayanan,
  ``Augmix: A simple data processing method to improve robustness and
  uncertainty,'' \emph{Proceedings of the International Conference on Learning
  Representations (ICLR)}, 2020.

\bibitem{Psaroudakis-cvpr2022}
A.~Psaroudakis and D.~Kollias, ``Mixaugment \& mixup: Augmentation methods for
  facial expression recognition,'' in \emph{2022 IEEE/CVF Conference on
  Computer Vision and Pattern Recognition Workshops (CVPRW)}.\hskip 1em plus
  0.5em minus 0.4em\relax IEEE, Jun. 2022, pp. 2366–--2374, doi:
  10.1109/cvprw56347.2022.00264.

\bibitem{LichaoSun-iccl2020}
L.~Sun, C.~Xia, W.~Yin, T.~Liang, P.~Yu, and L.~He, ``Mixup-transformer:
  Dynamic data augmentation for nlp tasks,'' in \emph{Proceedings of the 28th
  International Conference on Computational Linguistics}.\hskip 1em plus 0.5em
  minus 0.4em\relax International Committee on Computational Linguistics, 2020,
  doi: 10.18653/v1/2020.coling-main.305.

\bibitem{YunZhou-ml4its2023}
Y.~Zhou, L.~You, W.~Zhu, and P.~Xu, ``Improving time series forecasting with
  mixup data augmentation,'' in \emph{2nd International Workshop on Machine
  Learning for Irregular Time Series (ML4ITS2023)}, ser. -, vol.~-.\hskip 1em
  plus 0.5em minus 0.4em\relax -, Sept 2023, pp.~--,
  https://www.amazon.science/publications/improving-time-series-forecasting-with-mixup-data-augmentation.

\bibitem{Carratino-jmlr2022}
\BIBentryALTinterwordspacing
L.~Carratino, M.~Ciss\'{e}, R.~Jenatton, and J.-P. Vert, ``On mixup
  regularization,'' \emph{Journal of Machine Learning Research}, vol.~23, no.
  325, pp. 1--31, 2022. [Online]. Available:
  \url{http://jmlr.org/papers/v23/20-1385.html}
\BIBentrySTDinterwordspacing

\bibitem{YingtianZou-uai2023}
Y.~Zou, V.~Verma, S.~Mittal, W.~H. Tang, H.~Pham, J.~Kannala, Y.~Bengio,
  A.~Solin, and K.~Kawaguchi, ``{MixupE}: Understanding and improving mixup
  from directional derivative perspective,'' in \emph{Proceedings of the
  Thirty-Ninth Conference on Uncertainty in Artificial Intelligence}, ser.
  Proceedings of Machine Learning Research, R.~J. Evans and I.~Shpitser, Eds.,
  vol. 216.\hskip 1em plus 0.5em minus 0.4em\relax PMLR, 31 Jul--04 Aug 2023,
  pp. 2597--2607.

\bibitem{LinjunZhang-arxiv2021}
L.~Zhang, Z.~Deng, K.~Kawaguchi, A.~Ghorbani, and J.~Zou, ``How does mixup help
  with robustness and generalization?'' \emph{arXiv: 2010.04819}, 2021.

\bibitem{Thulasidasan-neurips2019}
S.~Thulasidasan, G.~Chennupati, J.~A. Bilmes, T.~Bhattacharya, and S.~Michalak,
  ``On mixup training: Improved calibration and predictive uncertainty for deep
  neural networks,'' in \emph{Advances in Neural Information Processing
  Systems, vol 32}, H.~Wallach, H.~Larochelle, A.~Beygelzimer, E.~Fox, and
  R.~Garnett, Eds.\hskip 1em plus 0.5em minus 0.4em\relax Curran Associates,
  Inc., 2019, pp.~--.

\bibitem{Zishanshao-arxiv2024}
Z.~Shao and A.~Devarakonda, ``Scalable dual coordinate descent for kernel
  methods,'' \emph{arXiv: 2406.18001}, 2024.

\bibitem{QiLei-pmlr17}
Q.~Lei, I.~E.-H. Yen, C.~yuan Wu, I.~S. Dhillon, and P.~Ravikumar, ``Doubly
  greedy primal-dual coordinate descent for sparse empirical risk
  minimization,'' in \emph{Proceedings of the 34th International Conference on
  Machine Learning}, ser. Proceedings of Machine Learning Research, D.~Precup
  and Y.~W. Teh, Eds., vol.~70.\hskip 1em plus 0.5em minus 0.4em\relax PMLR,
  06--11 Aug 2017, pp. 2034--2042.

\bibitem{DejunChu-tnn18}
D.~Chu, R.~Lu, J.~Li, X.~Yu, C.~Zhang, and Q.~Tao, ``Optimizing top-$k$
  multiclass {SVM} via semismooth newton algorithm,'' \emph{{IEEE} Transactions
  on Neural Networks and Learning Systems}, vol.~29, no.~12, pp. 6264--6275,
  Dec. 2018.

\bibitem{Tran-kdd2015}
K.~Tran, S.~Hosseini, L.~Xiao, T.~Finley, and M.~Bilenko, ``Scaling up
  stochastic dual coordinate ascent,'' in \emph{Proceedings of the 21th ACM
  SIGKDD International Conference on Knowledge Discovery and Data Mining}, ser.
  KDD' 15.\hskip 1em plus 0.5em minus 0.4em\relax ACM, Aug. 2015, doi:
  10.1145/2783258.2783412.

\bibitem{Takada-ieice24a}
Y.~Takada, R.~Mochida, M.~Nakajima, S.~suke Kadoya, D.~Sano, and T.~Kato,
  ``Stochastic dual coordinate ascent for learning sign constrained linear
  predictors,'' \emph{IEICE Transactions on Information and Systems}, vol.
  E107.D, no.~12, pp. 1493–--1503, Dec. 2024,
  doi:10.1587/transinf.2023edp7139.

\bibitem{TajTsu-icpr21a}
K.~Tajima, K.~Tsuchida, E.~R.~R. Zara, N.~Ohta, and T.~Kato, ``Learning
  sign-constrained support vector machines,'' in \emph{2020 25th International
  Conference on Pattern Recognition ({ICPR})}.\hskip 1em plus 0.5em minus
  0.4em\relax {IEEE}, Jan. 2021, doi:10.1109/icpr48806.2021.9412786.

\bibitem{TajHir-sdm21a}
K.~Tajima, Y.~Hirohashi, E.~R.~R. Zara, and T.~Kato, ``Frank-wolfe algorithm
  for learning svm-type multi-category classifiers,'' in \emph{Proceedings of
  SIAM International Conference on Data Mining (SDM21)}.\hskip 1em plus 0.5em
  minus 0.4em\relax Virginia, USA: SIAM, April 2021, pp.~--.

\bibitem{KatHir-acml19a}
T.~Kato and Y.~Hirohashi, ``Learning weighted top-$k$ support vector machine,''
  in \emph{Proceedings of The Eleventh Asian Conference on Machine Learning},
  ser. Proceedings of Machine Learning Research, W.~S. Lee and T.~Suzuki, Eds.,
  vol. 101.\hskip 1em plus 0.5em minus 0.4em\relax Nagoya, Japan: PMLR, 17--19
  Nov 2019, pp. 774--789.

\bibitem{Urruty01a}
J.-B. Hiriart-Urruty, \emph{Fundamentals of Convex Analysis}.\hskip 1em plus
  0.5em minus 0.4em\relax Springer, 2001.

\bibitem{StrombergThomas1996}
S.~Thomas, \emph{The operation of infimal convolution}.\hskip 1em plus 0.5em
  minus 0.4em\relax Instytut Matematyczny Polskiej Akademi Nauk, 1996.

\bibitem{YamAbuIma16a}
J.~Yamane, S.~Aburatani, S.~Imanishi, H.~Akanuma, R.~Nagano, T.~Kato, H.~Sone,
  S.~Ohsako, and W.~Fujibuchi, ``Prediction of developmental chemical toxicity
  based on gene networks of human embryonic stem cells,'' \emph{Nucleic Acids
  Research}, vol.~44, no.~12, pp. 5515--5528, July 2016.

\bibitem{Yamane-iScience22a}
J.~Yamane, T.~Wada, H.~Otsuki, K.~Inomata, M.~Suzuki, T.~Hisaki, S.~Sekine,
  H.~Kouzuki, K.~Kobayashi, H.~Sone, J.~K. Yamashita, M.~Osawa, M.~K. Saito,
  and W.~Fujibuchi, ``{StemPanTox}: A fast and wide-target drug assessment
  system for tailor-made safety evaluations using personalized {iPS} cells,''
  \emph{{iScience}}, vol.~25, no.~7, p. 104538, Jul. 2022,
  doi:10.1016/j.isci.2022.104538.

\bibitem{rockafellar70convex}
R.~T. Rockafellar, \emph{Convex Analysis}.\hskip 1em plus 0.5em minus
  0.4em\relax Princeton, NJ: Princeton University Press, 1970.

\end{thebibliography}

\appendices
\section{Smooth loss functions}
\label{s:stdloss}
This section gives some examples of the loss functions~$\phi_{0}$. 
The binary cross-entropy (BCE) loss 
\begin{tsaligned}
  \phi_{\text{bce}}(s):=\log(1+\exp(-s)), 
\end{tsaligned}
is an example of the function $\phi_{0}$. 
This loss function is calculated based on the log of the sigmoid function. 
The smooth hinge loss function
\begin{tsaligned}
    \phi_{\text{smh}}(s) :=
    \begin{cases}
      -s+1-\frac{\gamma_{\text{sm}}}{2} & \quad\text{if }s < 1 - \gamma_{\text{sm}}, \\
      \frac{1}{2\gamma_{\text{sm}}}(s-1)^{2} & \quad\text{if } 1 - \gamma_{\text{sm}} \le s < 1, \\
      0 & \quad\text{if } 1 \le s, 
    \end{cases}
\end{tsaligned}
where $\gamma_{\text{sm}}\in(0,1)$ is a constant, is a variant of the hinge loss function, which smooths the region where the margin is violated. It is commonly used in SVM. 
The quadratic hinge loss 
\begin{tsaligned}
  \phi_{\text{qh}}(s)
  :=
  \frac{1}{2\gamma_{\text{sm}}}\max(0,1-s)^{2} 
\end{tsaligned}
is another variant of the hinge loss function that penalizes margin violations more gradually.
The next section introduces the dual function, which includes the convex conjugate of the loss function. The convex conjugate of a function $\phi:\bR\to\bR$ is defined as:
\begin{tsaligned}\label{eq:conj-def}
  \phi^{*}(a)
  :=
  \sup_{\zeta\in\bR}
  a\zeta - \phi(\zeta). 
\end{tsaligned}

We assume that
\begin{tsaligned}
\phi_{0}(0) > 0, \, 
\forall s \le 0, \,
\nabla\phi_{0}(s) < 0. 
\end{tsaligned}  
Furthermore, assume that 
the conjugate $\phi_{0}^{*}$ is twice differentiable, 
and there exists $\gamma_{\text{sm}}>0$ such that
\begin{tsaligned}
    \forall a\in\text{dom}(\phi_{0}^{*}), 
    \qquad
    \nabla^{2}\phi_{0}^{*}(a)\ge \gamma_{\text{sm}}
\end{tsaligned}  
where $\text{dom}(f)$ is the effective domain of a function~$f$~\cite{rockafellar70convex}. 
The original function~$\phi_{0}$ is said to be a $1/\gamma_{\text{sm}}$-smooth function~\cite{Urruty01a}. Then, it is a well-known fact that 
$\forall \eta\in[0,1]$, $\forall u, \alpha \in -\text{dom}(\phi_{0}^{*})$, 
\begin{tsaligned}
  & \eta \phi_{0}^{*}(-u)
  +
  (1-\eta) \phi_{0}^{*}(-\alpha)
  \\
  & \ge
  \phi_{0}^{*}(-\eta u - (1-\eta)\alpha )
  +
  \frac{\gamma_{\text{sm}}}{2}(u-\alpha)^{2}(1-\eta)\eta. 
\end{tsaligned}
\newcommand{\tsnobar}[1]{{#1}}

\section{Proof for Lemma~\ref{lem:errp-if-geodecr}}

The assumption~\eqref{eq:geodecr} leads to the bound of the expected primal objective error with respect to randomness at previous iterations: 
\begin{tsaligned}
    \bE[h_{\text{P}}^{(t)}]
    &\le
    \tsnobar{\beta}^{-1}
    \bE\left[
    h_{\text{D}}^{(t)}-h_{\text{D}}^{(t+1)}
    \right]
    \le
    \tsnobar{\beta}^{-1}
    \bE\left[
    h_{\text{D}}^{(t)}
    \right]
    \\
    &
    \le
    \tsnobar{\beta}^{-1}
    \bE\left[
    h_{\text{D}}^{(t-1)}
    \right]
    (1-\tsnobar{\beta})
    \\
    &\le
    \tsnobar{\beta}^{-1}
    h_{\text{D}}^{(0)}
    \cdot(1-\tsnobar{\beta})^{t}
    \le
    \tsnobar{\beta}^{-1}
    h_{\text{D}}^{(0)}
    \exp\left(-\tsnobar{\beta}t\right). 
\end{tsaligned}
Hence, it holds that 
$\bE[h_{\text{P}}^{(t)}]\le\epsilon_{\text{P}}$ conditioned on
\(
    \tsnobar{\beta}^{-1}
h_{\text{D}}^{(0)}
\exp\left(-\tsnobar{\beta}t\right)
\le
\epsilon_{\text{P}}
\). 
This condition can be rearranged as
\begin{tsaligned}
t 
\ge 
\frac{1}{\tsnobar{\beta}}
\log\left(
  \frac{h_{\text{D}}^{(0)}}{\epsilon_{\text{P}}\tsnobar{\beta}}
\right). 
\end{tsaligned}
\tsqed

\section{Proof for Theorem~\ref{thm:approx-beta}}
We shall first show that 
\begin{tsaligned}\label{eq:J2-le-J2}
    {J}_{t}^{\text{van}}(\bar{s}_{t})
    \le
    {J}_{t}^{\text{van}}(\eta_{\text{approx},t}). 
\end{tsaligned}
Let
\begin{tsaligned}
  \widehat{\eta}_{\text{II}}
  :=
\bar{s}_{t}
\cdot
\frac{F_{\text{van},i}^{(t-1)}+\frac{1}{2}\gamma_{\text{sm}} q_{i,t}^{2}}{\gamma_{\text{sm}} q_{i,t}^{2}}
\end{tsaligned}
and
\begin{tsaligned}
  \widetilde{\eta}_{0}
:=
\bar{s}_{t}
\cdot
\frac{\widetilde{F}_{t}+\frac{1}{2}\gamma_{\text{sm}} q_{i,t}^{2}}{\gamma_{\text{sm}} q_{i,t}^{2}}. 
\end{tsaligned}
From Lemma~\ref{lem:tilF-le-F}, 
\begin{tsaligned}
  \widetilde{\eta}_{0} \le \widehat{\eta}_{\text{II}}. 
\end{tsaligned}
Function $J^{\text{van}}_{t}:\bR\to\bR$ is increasing in the interval $(-\infty,\widehat{\eta}_{\text{II}})$
and decreasing in the interval $(\widehat{\eta}_{\text{II}},+\infty)$, 
since the derivative of $J^{\text{van}}_{t}$ is given by 
\begin{tsaligned}
  \nabla J^{\text{van}}_{t}(\eta)
  =
  \frac{\gamma_{\text{sm}}q_{i,t}^{2} }{n \bar{s}_{t} } 
  \left(
    \widehat{\eta}_{\text{II}}
    % \bar{s}_{t} \cdot
    % \frac{%
    % F_{\text{van},i}^{(t-1)}
    % +
    % \frac{1}{2}\gamma_{\text{sm}}q_{i}^{2}
    % }{
    %   \gamma_{\text{sm}}q_{i}^{2}
    % }
    -
    \eta
      \right). 
\end{tsaligned}
There are two cases: 
i) $\widetilde{\eta}_{0} \le \bar{s}_{t}$, 
ii) $\bar{s}_{t} < \widetilde{\eta}_{0}$. 
% ii) $\bar{s}_{t} < \widetilde{\eta}_{0} \le \widehat{\eta}_{\text{II}}$. 
% iii) $\widehat{\eta}_{\text{II}} < \widetilde{\eta}_{0}$. 
%
\begin{itemize}
\item
In case of $\widetilde{\eta}_{0} \le \bar{s}_{t}$, we have $\eta_{\text{approx},t}=\bar{s}_{t}$, 
thereby 
\begin{tsaligned}
  {J}_{t}^{\text{van}}(\eta_{\text{approx},t})
  =
  {J}_{t}^{\text{van}}(\bar{s}_{t}). 
\end{tsaligned}
\item
In case of $\bar{s}_{t} < \widetilde{\eta}_{0}$, we have 
\begin{tsaligned}
  \bar{s}_{t} < 
  \eta_{\text{approx},t}
  =
  \min
  \left\{
    1,
    \widetilde{\eta}_{0}
  \right\}
  \le
  \widehat{\eta}_{\text{II}}. 
\end{tsaligned}
Recall that 
the function $J^{2}_{t}:\bR\to\bR$ is increasing in the interval $(-\infty,\widehat{\eta}_{\text{II}})$. 
Therefore, we get
\begin{tsaligned}
    {J}_{t}^{\text{van}}(\bar{s}_{t})
    \le
    {J}_{t}^{\text{van}}(\eta_{\text{approx},t}). 
  \end{tsaligned}
\end{itemize}
Thus, the inequality~\eqref{eq:J2-le-J2} has been established.  

We next observe that 
\begin{tsaligned}
    \lVert f_{0,\valph^{(t-1)}}\rVert^{2}  
    &=
    \left<
        f_{0,\valph^{(t-1)}}, 
    \frac{1}{\lambda n}
    \sum_{j=1}^{n}
    \alpha_{j}^{(t-1)}
    \kappa(\x_{j},\cdot)
    \right>
    \\
    &=
    \frac{1}{\lambda n}    
    \sum_{j=1}^{n}
    z_{j}^{(t-1)}\alpha_{j}^{(t-1)}    
\end{tsaligned}
leading to
\begin{tsaligned}\label{eq:dualgap-ge-F-ave}
    & 
    h_{\text{P}}^{(t-1)} + h_{\text{D}}^{(t-1)} 
    =
    R[f_{0,\valph^{(t-1)}}] - D_{0}(\valph^{(t-1)})
    \\
    &=
    \frac{1}{n}    
    \sum_{j=1}^{n}
    \phi_{\text{mup}}(z_{j}^{(t-1)}\,;\,y_{j}) 
    +
    \phi^{*}_{\text{mup}}(-\alpha_{j}^{(t-1)}\,;\,y_{j}) 
    \\
    &
    +
    \lambda\lVert f_{0,\valph^{(t-1)}}\rVert^{2}  
    =
    \frac{1}{n}    
    \sum_{j=1}^{n}
    F_{j}^{(t-1)}. 
\end{tsaligned}
We now use the inequality~\eqref{eq:J2-le-J2} to have
\begin{tsaligned}\label{eq:dual-impr-ge-F-i}
    &
    h_{\text{D}}^{(t-1)}-h_{\text{D}}^{(t)}
    =
    J^{0}_{t}(q_{i,t}\eta_{\text{approx},t})
    \ge 
    {J}_{t}^{\text{van}}(\eta_{\text{approx},t})
    \\
    &
    \ge
    {J}_{t}^{\text{van}}(\bar{s}_{t})
    =
    \frac{\bar{s}_{t}}{n}
    F_{\text{van},i}^{(t-1)}. 
\end{tsaligned}
Taking the expectation with respect to the randomness for selection of $i\in[n]$ at $t$th iteration, we obtain
\begin{tsaligned}\label{eq:ssslem19}
    &
    h_{\text{D}}^{(t-1)}-\bE\left[h_{\text{D}}^{(t)}\right]
    \ge
    \frac{1}{n}
    \frac{\lambda n \gamma_{\text{sm}}}{R^{2}_{\text{mx}} + \lambda n\gamma_{\text{sm}}}    
    \sum_{i=1}^{n}F_{\text{van},i}^{(t-1)}
    \\
    &
    \ge
    \frac{\lambda \gamma_{\text{sm}}}{R^{2}_{\text{mx}} + \lambda n\gamma_{\text{sm}}}    
    \cdot\left(h_{\text{P}}^{(t-1)} + h_{\text{D}}^{(t-1)} \right)
    \\
    &
    \ge
    \frac{1}{n + R^{2}_{\text{mx}}/(\lambda \gamma_{\text{sm}})}
    \cdot
    \max\left\{
        h_{\text{P}}^{(t-1)}, h_{\text{D}}^{(t-1)}
    \right\}. 
\end{tsaligned}
This implies that \eqref{eq:geodecr} is satisfied with 
\begin{tsaligned}
  \beta^{-1} = n + \frac{R^{2}_{\text{mx}}}{\lambda \gamma_{\text{sm}}}. 
\end{tsaligned}
which allows us to apply Lemma~\ref{lem:errp-if-geodecr}. 
Hence, Theorem~\ref{thm:approx-beta} is established. 
\tsqed

\section{Proof for Lemma~\ref{lem:tilphi-smooth}}
The convex conjugate of $\widetilde{\phi}_{i}$ is expressed as
\begin{tsaligned}
  \widetilde{\phi}^{*}_{i}(a)
  &=
  \sup_{s\in\bR}
  \left( sa - \widetilde{\phi}_{i}(s) \right)
  \\
  &=
  \sup_{s\in\bR}
  \left( sa - (1+y_{i})\phi_{0}(s) \right)
  \\
  &=
  (1+y_{i})
  \sup_{s\in\bR}
  \left( \frac{sa}{1+y_{i}} - \phi_{0}(s) \right)
  \\
  &=
  (1+y_{i})\phi^{*}_{0}
  \left(\frac{a}{1+y_{i}}\right). 
\end{tsaligned}
Recall the assumption that the value of $\phi^{*}_{0}$ can be computed with $O(1)$ cost, and so is that of $\widetilde{\phi}^{*}_{i}$. 

The second order derivative is: 
\begin{tsaligned}
  \nabla^{2}\widetilde{\phi}^{*}_{i}(a)
  =
  \frac{1}{1+y_{i}}
  \nabla^{2}\phi_{0}^{*}
  \left(\frac{a}{1+y_{i}}\right)
  \ge
  \frac{\gamma_{\text{sm}}}{1+y_{i}}
  \ge 
  \frac{\gamma_{\text{sm}}}{2} 
\end{tsaligned}
which implies that $\widetilde{\phi}_{i}$ is a $2/\gamma_{\text{sm}}$-smooth function. 
\tsqed

\section{Proof for Theorem~\ref{thm:decomp-beta}}
Let
\begin{tsaligned}
  \widetilde{h}_{\text{D}}^{(t)}
  :=
  \widetilde{D}(\valph_{\star})
  -
  \widetilde{D}(\valph^{(t)}). 
\end{tsaligned}
Using the similar proof technique used in Theorem~\ref{thm:approx-beta}, 
we have 
\begin{tsaligned}\label{eq:ssslem19}
    &
    \widetilde{h}_{\text{D}}^{(t-1)}-\bE\left[\widetilde{h}_{\text{D}}^{(t)}\right]
    \ge
    \frac{\lambda \gamma_{\text{sm}}/2}{R^{2}_{\text{mx}} + \lambda \widetilde{n}\gamma_{\text{sm}}/2}
    \cdot
    \left(h_{\text{P}}^{(t-1)} + \widetilde{h}_{\text{D}}^{(t-1)} \right)
    \\
    &
    \ge
    \frac{1}{2n + 2R^{2}_{\text{mx}}/(\lambda \gamma_{\text{sm}})}
        \cdot
    \max\left\{
        h_{\text{P}}^{(t-1)}, \widetilde{h}_{\text{D}}^{(t-1)}
    \right\} 
\end{tsaligned}
implying the assumption in \eqref{eq:geodecr} is satisfied with 
\begin{tsaligned}
  \beta^{-1} = 2n + \frac{2R^{2}_{\text{mx}}}{\lambda \gamma_{\text{sm}}}. 
\end{tsaligned}
Applying Lemma~\ref{lem:errp-if-geodecr} concludes the proof. 
\tsqed

\end{document}